\documentclass[10pt,twocolumn,letterpaper]{article}

\usepackage{iccv}
\usepackage{times}
\usepackage{epsfig}
\usepackage{graphicx}
\usepackage{amsmath}
\usepackage{amssymb}
\usepackage{booktabs}
\usepackage{subfigure}
\usepackage{caption}
\usepackage{array}
\usepackage{tikz}
\usepackage{pdfpages}

\usepackage{color}
\usepackage[breaklinks=true,bookmarks=false]{hyperref}

\begin{document}

\iccvfinalcopy % *** Uncomment this line for the final submission

\def\httilde{\mbox{\tt\raisebxox{-.5ex}{\symbol{126}}}}

% Pages are numbered in submission mode, and unnumbered in camera-ready
% \ificcvfinal\pagestyle{empty}\fi
% \setcounter{page}{1}

% Example definitions.
% --------------------
\def\x{{\boldsymbol x}}
\def\y{{\boldsymbol y}}
\def\c{{\boldsymbol c}}
\def\A{{\boldsymbol A}}
\def\Q{{\boldsymbol Q}}
\def\K{{\boldsymbol K}}
\def\V{{\boldsymbol V}}
\def\L{{\cal L}}

\definecolor{my_red}{RGB}{237, 2, 140}

\title{Attention on Attention for Image Captioning}

\author{
  Lun Huang$^{1}$ \quad Wenmin Wang$^{1,3}$\thanks{Corresponding author} \quad Jie Chen$^{1,2}$   \quad Xiao-Yong Wei$^{2}$ \\
  $^{1}$School of Electronic and Computer Engineering, Peking University\\
  % $^{2}$Peng Cheng Laboratory \quad $^{3}$Macau University of Science and Technology\\
  $^{2}$Peng Cheng Laboratory \\
  $^{3}$Macau University of Science and Technology\\
{\tt\small huanglun@pku.edu.cn, \{wangwm@ece.pku.edu.cn, wmwang@must.edu.mo\}, \{chenj, weixy\}@pcl.ac.cn}}

\maketitle

% \ificcvfinal\thispagestyle{empty}\fi
\begin{abstract}
Attention mechanisms are widely used in current encoder/decoder frameworks of image captioning, where a weighted average on encoded vectors is generated at each time step to guide the caption decoding process. However, the decoder has little idea of whether or how well the attended vector and the given attention query are related, which could make the decoder give misled results. In this paper, we propose an ``Attention on Attention'' (AoA) module, which extends the conventional attention mechanisms to determine the relevance between attention results and queries. AoA first generates an ``\textbf{information vector}'' and an ``\textbf{attention gate}'' using the attention result and the current context, then adds another attention by applying element-wise multiplication to them and finally obtains the ``\textbf{attended information}'', the expected useful knowledge. We apply AoA to both the encoder and the decoder of our image captioning model, which we name as AoA Network (AoANet). Experiments show that AoANet outperforms all previously published methods and achieves a new state-of-the-art performance of 129.8 CIDEr-D score on MS COCO ``Karpathy'' offline test split and 129.6 CIDEr-D (C40) score on the official online testing server. Code is available at \url{https://github.com/husthuaan/AoANet}.
\end{abstract}

\section{Introduction}
\label{sec:intro}

\begin{figure}[t]
	\begin{center}
		\includegraphics[width=0.98\linewidth]{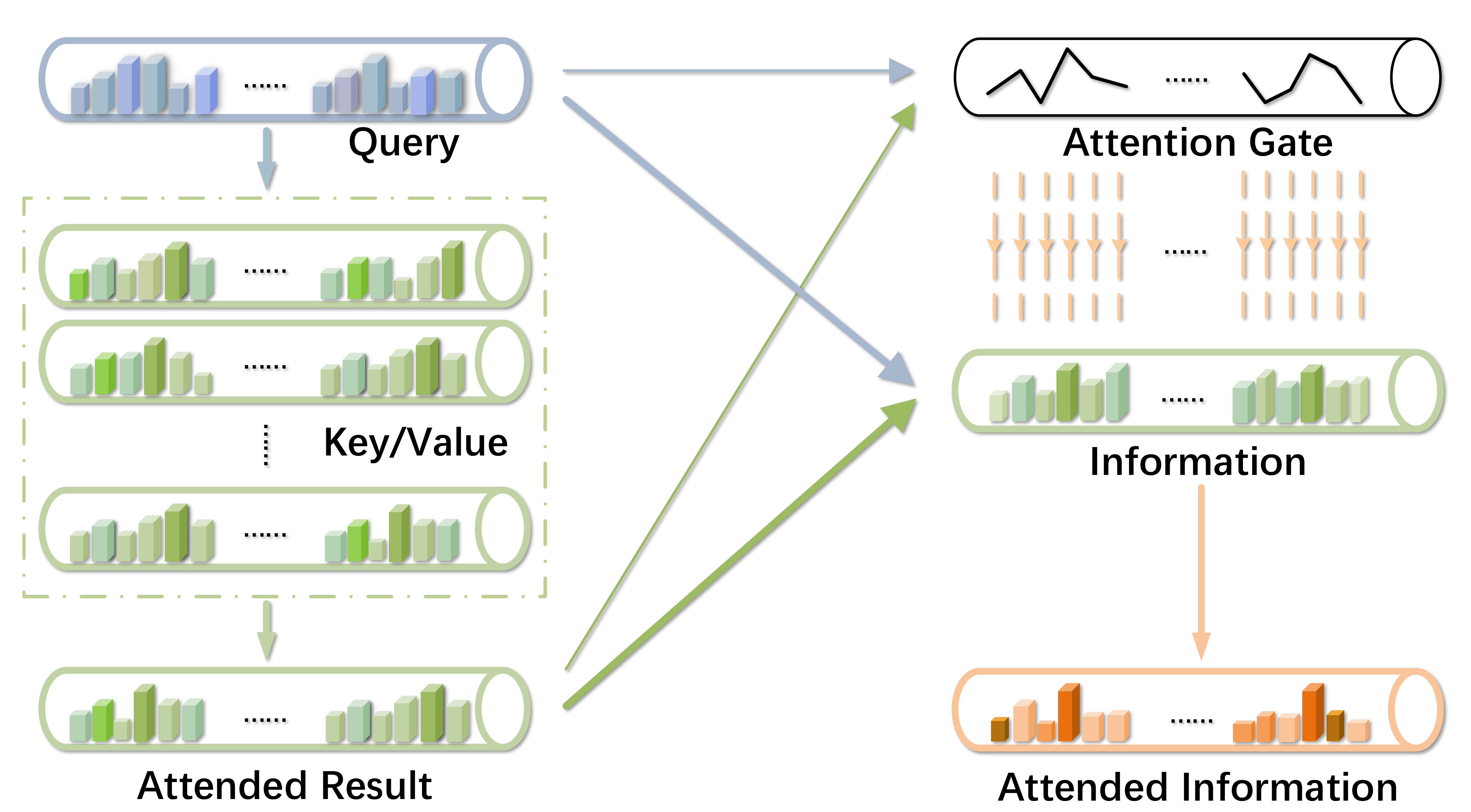}
	\end{center}
	\caption{\emph{Attention on Attention} (AoA). AoA generates an \textbf{information vector} and an \textbf{attention gate} using the attention result and the attention query, and adds another attention by applying the gate to the information and obtains the \textbf{attended information}.}
	\label{fig:overall}
\end{figure}

Image captioning is one of the primary goals of computer vision which aims to automatically generate natural descriptions for images. It requires not only to recognize salient objects in an image,  understand their interactions, but also to verbalize them using natural language, which makes itself very challenging \cite{kulkarni2013babytalk,yang2011corpus,mitchell2012midge,fang2015captions}.

Inspired by the development of neural machine translation, attention mechanisms have been widely used in current encoder/decoder frameworks for visual captioning \cite{Xu2015Show, Lu2017Knowing, Yang2016Review, Anderson2018Bottom, chen2018less,gan2017semantic,Gan_2017_CVPR} and achieved impressive results.
In such a framework for image captioning, an image is first encoded to a set of feature vectors via a CNN based network and then decoded to words via an RNN based network,
where the attention mechanism guides the decoding process by generating a weighted average over the extracted feature vectors for each time step.

The attention mechanism plays a crucial role in such a system that must capture global dependencies, 
\eg a model for the sequence to sequence learning task like image/video captioning, 
since the output is directly conditioned on the attention result.
However, the decoder has little idea of whether or how well the attention result is related to the query.
There are some cases when the attention result is not what the decoder expects and the decoder can be misled to give fallacious results, which could happen when the attention module doesn't do well on its part or there's no worthful information from the candidate vectors at all. The former case can't be avoided since mistakes always happen. As for the latter, when there's nothing that meets the requirement of a specific query, the attention module still returns a vector which is a weighted average on the candidate vectors and thus is totally irrelevant to the query.

To address this issue, we propose \emph{Attention on Attention} (AoA),
which extends the conventional attention mechanisms by adding another attention.
Firstly, AoA generates an ``\textbf{information vector}'' and an ``\textbf{attention gate}'' with two linear transformations, which is similar to GLU~\cite{dauphin2016language}. The information vector is derived from the current context (\ie the query) and the attention result via a linear transformation, and stores the newly obtained information from the attention result together with the information from the current context. The attention gate is also derived from the query and the attention result via another linear transformation with sigmoid activation followed, and the value of each channel indicates the relevance/importance of the information on the corresponding channel in the information vector.
Subsequently, AoA adds another attention by applying the attention gate to the information vector using element-wise multiplication and finally obtains the ``\textbf{attended information}'', the expected useful knowledge.

AoA can be applied to various attention mechanisms. For the traditional single-head attention, AoA helps to determine the relevance between the attention result and query. Specially, for the recently proposed multi-head attention \cite{vaswani2017attention}, AoA helps to build relationships among different attention heads, filters all the attention results and keeps only the useful ones.

We apply AoA to both the image encoder and the caption decoder of our image captioning model, AoANet.
For the encoder, it extracts feature vectors of objects in the image, applies self-attention \cite{vaswani2017attention} to the vectors to model relationships among the objects, and then applies AoA to determine how they are related to each other. For the decoder, it applies AoA to filter out the irrelevant/misleading attention results and keep only the useful ones.

We evaluate the impact of applying AoA to the encoder and decoder respectively. Both quantitative and qualitative results show that AoA module is effective.
The proposed AoANet outperforms all previously published image captioning models: a single model of AoANet achieves 129.8 CIDEr-D score on MS COCO dataset offline test split; and an ensemble of 4 models achieves 129.6 CIDEr-D (C40) score on the online testing server. Main contributions of this paper include:
\begin{itemize}
  \vspace{-0.1cm}
  \item
  We propose the \emph{Attention on Attention} (AoA) module, an extension to the conventional attention mechanism, to determine the relevance of attention results.
  \vspace{-0.1cm}
  \item
  We apply AoA to both the encoder and decoder to constitute AoANet: in the encoder, AoA helps to better model relationships among different objects in the image; in the decoder, AoA filters out irrelative attention results and keeps only the useful ones.
  \vspace{-0.1cm}
  \item
  % We set a new state-of-the-art performance on MS COCO dataset.
  Our method achieves a new state-of-the-art performance on MS COCO dataset.
\end{itemize}

\section{Related Work}

% AoA extends conventional attention mechanisms to determine the relevance between the attention result and query, and is applied to image captioning in this paper.

\subsection{Image Captioning}
Earlier approaches to image captioning are rule/template-based \cite{yao2010i2t,socher2010connecting} which generate slotted caption templates and use the outputs of object detection \cite{Ren2015Faster,wan2019min-entropy,Wan_2019_CVPR}, attribute prediction and scene recognition to fill in the slots.
Recent approaches are neural-based and specifically, utilize a deep encoder decoder framework, which is inspired by the development of neural machine translation \cite{cho2014learning}.
For instance, an end-to-end framework is proposed with a CNN encoding the image to feature vector and an LSTM decoding it to caption \cite{Vinyals2015Show}.
In \cite{Xu2015Show}, the spatial attention mechanism on CNN feature map is used to incorporate visual context. In \cite{Chen_2017_CVPR}, a spatial and channel-wise attention model is proposed.
In \cite{Lu2017Knowing}, an adaptive attention mechanism is introduced to decide when to activate the visual attention. More recently, more complex information such as objects, attributes and relationships are integrated to generate better descriptions \cite{yao2017boosting,Anderson2018Bottom,yao2018exploring,yang2019auto}.

\subsection{Attention Mechanisms}
The attention mechanism \cite{rensink2000the,corbetta2002control}, which is derived from human intuition, has been widely applied and yielded significant improvements for various sequence learning tasks.
It first calculates an importance score for each candidate vector, then normalizes the scores to weights using the soft-max function, finally applies these weights to the candidates to generate the attention result, a weighted average vector \cite{Xu2015Show}. There are other attention mechanisms such as: spatial and channel-wise attention \cite{Chen_2017_CVPR}, adaptive attention \cite{Lu2017Knowing}, stacked attention \cite{Yang_2016_CVPR}, multi-level attention \cite{Yu_2017_CVPR}, multi-head attention and self-attention \cite{vaswani2017attention}.

Recently, Vaswani et al. \cite{vaswani2017attention} showed that solely using self-attention can achieve state-of-the-art results for machine translation.
Several works extend the idea of employing self-attention to some tasks \cite{wang2018non-local,hu2018relation} in computer vision, which inspires us to apply self-attention to image captioning to model relationships among objects in an image.

\subsection{Other Work}
AoA generates an attention gate and an information vector via two linear transformations and applies the gate to the vector to add a second attention, where the techniques are similar to some other work:
GLU~\cite{dauphin2016language}, which replaces RNN and CNN to capture long-range dependencies for language modeling; multi-modal fusion~\cite{Yang2015Compact,Fukui2016Multimodal,Kim2017Hadamard,Benyounes_2017_ICCV,Gao_2019_CVPR}, which models interactions between different modalities (\eg text and image) and combines information from them; LSTM/GRU, which uses gates and memories to model its inputs in a sequential manner.

\subsection{Summarization}
% AoA extends conventional attention mechanisms to determine the relevance between the attention result and query, and is applied to image captioning in this paper.
We summarize the differences between our method and the work discussed above, as follows: We apply \emph{Attention on Attention} (AoA) to image captioning in this paper; AoA is a general extension to attention mechanisms and can be applied to any of them; AoA determines the relevance between the attention result and query, while multi-modal fusion combines information from different modalities; AoA requires only one ``attention gate'' but no hidden states. In contrast, LSTM/GRU requires hidden states and more gates, and is applicable only to sequence modeling.

\section{Method}
We first introduce the \emph{Attention on Attention} (AoA) module and then show how we derive AoANet for image captioning by applying AoA to the image encoder and the caption decoder.

\begin{figure}[t] 
  \subfigure[Attention]{  
    \begin{minipage}[t]{0.45\linewidth}
    \centering                                       
    \includegraphics[scale=0.6]{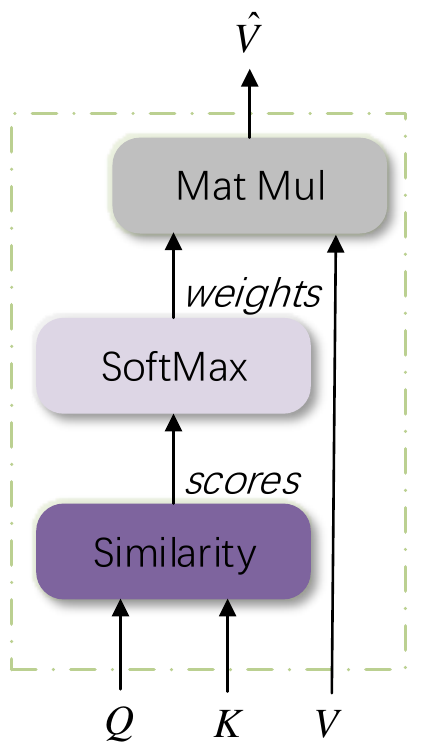}   
    \label{fig:aoa:attention}   
    \end{minipage}
  }
  \subfigure[Attention on Attention]{  
    \begin{minipage}[t]{0.48\linewidth}
    \centering                                       
    \includegraphics[scale=0.5]{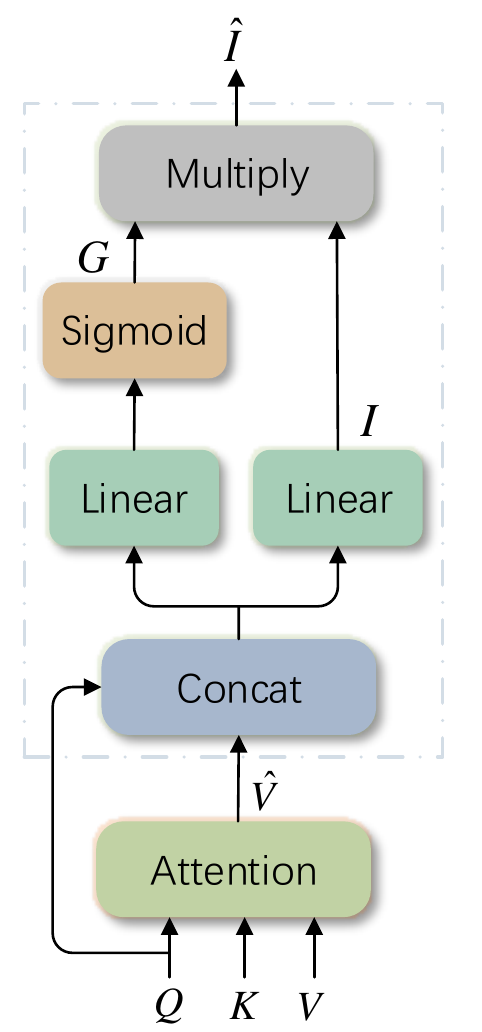}   
    \label{fig:aoa:aoa}    
    \end{minipage}
  }     
\caption{Attention and ``\emph{Attention on Attention}'' (AoA). \textbf{(a)}~The attention module generates some weighted average $\hat{\V}$ based on the similarity scores between $\Q$ and $\K$; \textbf{(b)}~AoA generates the ``information vector'' $\boldsymbol{I}$ and ``attention gate'' $\boldsymbol{G}$, and adds another attention via element-wise multiplication.}
\label{fig:aoa}         
\end{figure}

% \begin{figure}[!ht] 
%   \centering
%   \subfigure[Attention]{  
%     \parbox[][6cm][c]{\linewidth}{
%     \includegraphics[scale=0.1]{figs_embed/attention-module.pdf}  } 
%   } 
%   \subfigure[Attention]{  
%     \parbox[][6cm][c]{\linewidth}{
%     \includegraphics[scale=0.1]{figs_embed/aoa-module.pdf}  } 
%   } 
% \caption{Attention and ``Attention on Attention'' (AoA). \textbf{(a)} The attention module generates some weighted average $\hat{\V}$ based on the similarity scores between $\Q$ and $\K$; \textbf{(b)} AoA generates the ``information vector'' $\boldsymbol{I}$ and ``attention gate'' $\boldsymbol{G}$ and adds another attention via element-wise multiplication.}
% \label{fig:aoa}         
% \end{figure}

\subsection{Attention on Attention}
An attention module $f_{att}(\boldsymbol{Q,K,V})$ operates on some queries, keys and values and generates some weighted average vectors (denoted by $\Q$, $\K$, $\V$ and $\hat{\V}$ respectively), in Figure \ref{fig:aoa:attention}. It first measures the similarities between $\Q$ and $\K$ and then uses the similarity scores to compute weighted average vectors over $\V$, 
which can be formulated as:
\begin{align}
  a_{i,j} &= f_{sim}(\boldsymbol{q}_i, \boldsymbol{k}_j), \alpha_{i,j} =  \frac{e^{a_{i,j}}}{\sum_{j}{e^{a_{i,j}}}}\\
  \boldsymbol{\hat{v}_i} &= \sum_{j}{\alpha_{i,j} \boldsymbol{v}_j}
\end{align}
where $\boldsymbol{q}_i \in \Q$ is the $i^{th}$ query, $\boldsymbol{k}_j \in \boldsymbol{K}$ and $\boldsymbol{v}_j \in \boldsymbol{V}$ are the $j^{th}$ key/value pair; $f_{sim}$ is a function that computes the similarity score of each $\boldsymbol{k}_j$ and $\boldsymbol{q}_i$; and $\boldsymbol{\hat{v}}_i$ is the attended vector for the query $\boldsymbol{q}_i$.

The attention module outputs a weighted average for each query, no matter whether or how $\Q$ and $\K$/$\V$ are related. Even when there is no relevant vectors, the attention module still generates a weighted average vector, which can be irrelevant or even misleading information.

Thus we propose the AoA module (as shown in Figure \ref{fig:aoa:aoa}) to measure the relevance between the attention result and the query.
The AoA module generates an ``information vector'' $\boldsymbol{i}$ and an ``attention gate'' $\boldsymbol{g}$ via two separate linear transformations, which are both conditioned on the attention result and the current context (\ie the query) $\boldsymbol{q}$:
\begin{align}
  % i &= W^i_c \boldsymbol{c}_t + W^i_v \boldsymbol{\hat{v}} + b^i \\
  \boldsymbol{i} = W^i_q \boldsymbol{q} + W^i_v \boldsymbol{\hat{v}} + b^i
\end{align} 
\begin{align}
  \boldsymbol{g} = \sigma( W^g_q \boldsymbol{q} + W^g_v \boldsymbol{\hat{v}} + b^g)
\end{align}
where $W^i_q, W^i_v, W^g_q, W^g_v \in \mathbb{R}^{D \times D}$, $b^i, b^g \in \mathbb{R}^{D}$, and $D$ is the dimension of $\boldsymbol{q}$ and $\boldsymbol{v}$; $\boldsymbol{\hat{v}} = f_{att}(\boldsymbol{Q,K,V})$ is the attention result, $f_{att}$ is an attention module and $\sigma$ denotes the sigmoid activation function.

Then AoA adds another attention by applying the attention gate to the information vector using element-wise multiplication and obtains the attended information $\hat{\boldsymbol{i}}$:
\begin{align}
  \hat{\boldsymbol{i}} = \boldsymbol{g} \odot \boldsymbol{i}
\end{align}
where $\odot$ denotes element-wise multiplication. The throughout pipeline of AoA is formulated as:
\begin{multline}
    \textrm{AoA}(f_{att},\boldsymbol{Q,K,V}) = \sigma( W^g_q \Q + W^g_v f_{att}(\boldsymbol{Q,K,V}) + b^g) \\
     \odot (W^i_q \Q + W^i_v f_{att}(\boldsymbol{Q,K,V}) + b^i) 
\end{multline}

\subsection{AoANet for Image Captioning}
\begin{figure}[t]
	\begin{center}
		\includegraphics[width=0.98\linewidth]{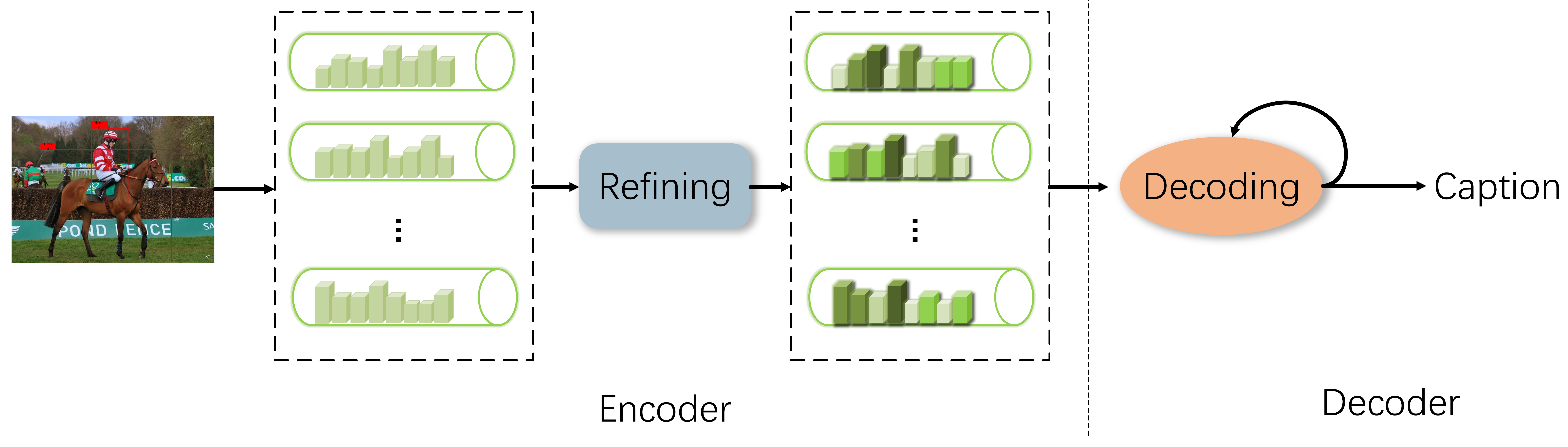}
	\end{center}
	\caption{Overview of the encoder/decoder framework of AoANet. A refining module is added in the encoder to model relationships of objects in the image.}
	\label{fig:aoanet}
\end{figure}

\begin{figure}[ht]
	\begin{center}
		\includegraphics[width=0.88\linewidth]{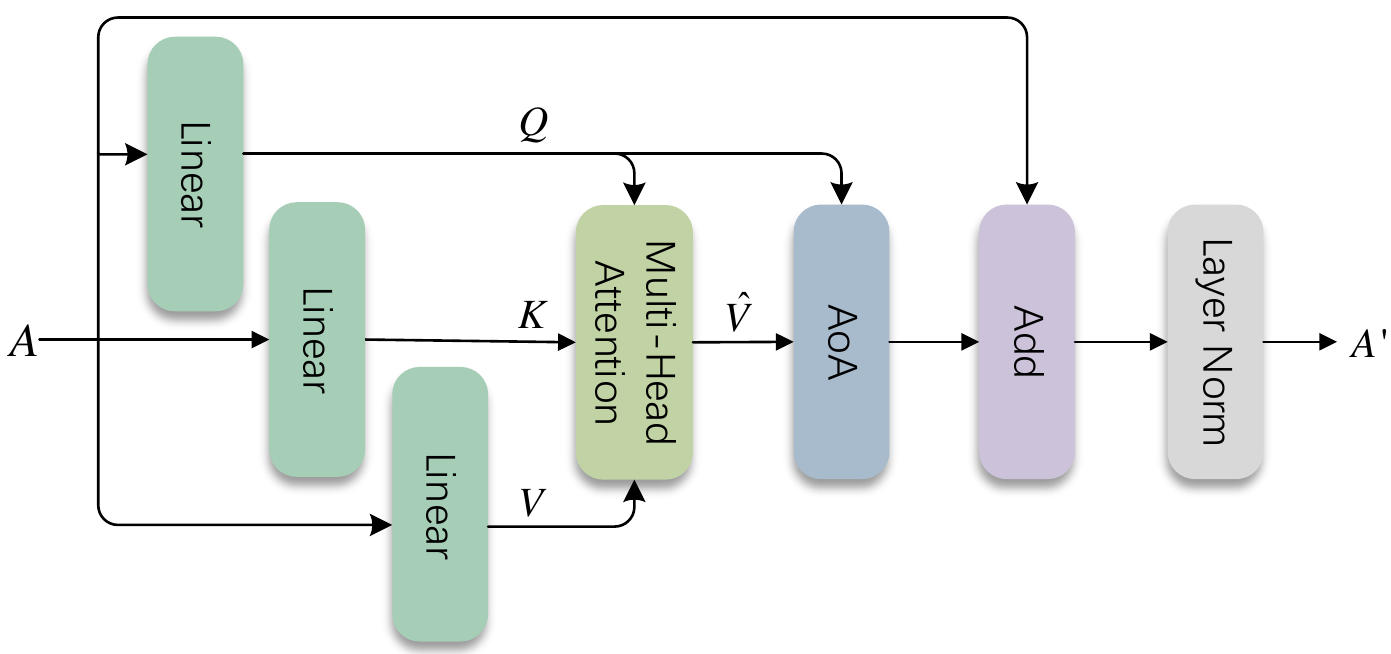}
	\end{center}
	\caption{The refining module in the image encoder, where AoA and the self-attentive multi-head attention refine the representations of feature vectors by modeling relationships among them.}
	\label{fig:encoder}
\end{figure}
\begin{figure}[t]
	\begin{center}
		\includegraphics[width=0.98\linewidth]{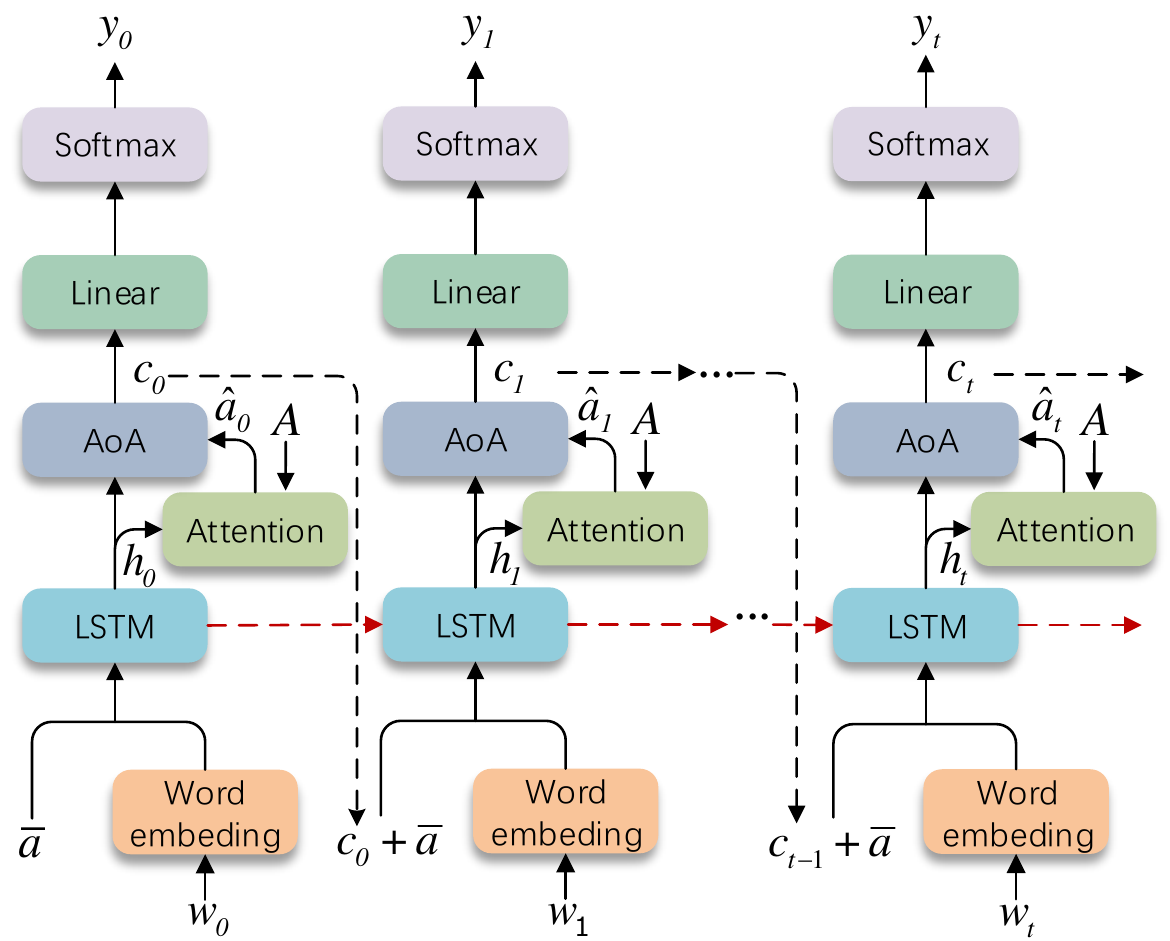}
	\end{center}
	\caption{The caption decoder of AoANet, which contains an LSTM, an AoA module and a word prediction module.}
	\label{fig:decoder}
\end{figure}

We build the model, AoANet, for image captioning based on the encoder/decoder framework (Figure \ref{fig:aoanet}), where both the encoder and the decoder are incorporated with an AoA module.

\subsubsection{Encoder with AoA}
For an image, we first extract a set of feature vectors $\boldsymbol{A} = \{\boldsymbol{a_1,a_2,...,a_k}\}$ using a CNN or R-CNN based network, where $\boldsymbol{a}_i \in \mathbb{R}^D$, $k$ is the number of vectors in $\A$, and $D$ is the dimension of each vector.

Instead of directly feeding these vectors to the decoder, we build a refining network which contains an AoA module to refine their representations (Figure \ref{fig:encoder}). The AoA module in the encoder, notated as $\textrm{AoA}^E$, adopts the multi-head attention function~\cite{vaswani2017attention} where $\Q, \K$, and $\V$ are three individual linear projections of the feature vectors $\A$.
The AoA module is followed by a residual connection~\cite{He_2016_CVPR} and layer normalization~\cite{Ba2016Layer}:
\begin{multline}
  \A' =  \textrm{LayerNorm}(\A + \\
   \textrm{AoA}^E(f_{mh-att}, W^{Q_e} \A, W^{K_e} \A, W^{V_e} \A))
\end{multline}
where $W^{Q_e}, W^{K_e}, W^{V_e} \in \mathbb{R}^{D \times D}$ are three linear transformation matrixes. $f_{mh-att}$ is the multi-head attention function which divides each $\Q, \K, \V$ into $H=8$ slices along the channel dimension, and employs a scaled dot-product attention function $f_{dot-att}$ to each slice $\Q_i, \K_i, \V_i$, then concatenates the results of each slice to form the final attended vector.

\begin{align}
  f_{mh-att}(\Q, \K, \V)= \textrm{Concat}(head_1,...,head_H)
\end{align}
\begin{align}
    head_i = f_{dot-att}(\Q_i , \K_i , \V_i)
\end{align}
\begin{align}
  f_{dot-att}(\Q_i,\K_i,\V_i) = \textrm{softmax}(\frac{\Q_i\K_i^T}{\sqrt{d}})\V_i
\end{align}

In this refining module, the self-attentive multi-head attention module seeks the interactions among objects in the image, and AoA is applied to measure how well they are related. After refining, we update the feature vectors $\A \leftarrow \A'$.
The refining module doesn't change the dimension of $\A$, and thus can be stacked for $N$ times ($N=6$ in this paper).

Note that the refining module adopts a different structure from that of the original transformer encoder~\cite{vaswani2017attention} as the feed-forward layer is dropped, which is optional and the change is made for the following two reasons:
1) the feed-forward layer is added to provide non-linear representations, which is also realized by applying AoA;
2) dropping the feed-forward layer does not change the performances perceptually of AoANet but gives simplicity.

\subsubsection{Decoder with AoA}
The decoder (Figure \ref{fig:decoder}) generates a sequence of caption $\boldsymbol{y}$ with the (refined) feature vectors $\A$.
% (we refer both the original feature vectors $\A$ and refined vectors $\A'$ to be $\A$ for convenience).

We model a context vector $\boldsymbol{c}_t$ to compute the conditional probabilities on the vocabulary:
\begin{align}
  p(\boldsymbol{y}_t \mid \boldsymbol{y}_{1:t-1}, I) = \textrm{softmax}(W_p \boldsymbol{c}_t)
\end{align}
where $W_p \in \mathbb{R}^{D \times |\Sigma|}$ is the weight parameters to be learnt and $|\Sigma|$ the size of the vocabulary.

\begin{table*}[t]
  \centering
  \label{tab:examples}
  \caption{Performance of our model and other state-of-the-art methods on MS-COCO ``Karpathy'' test split, where B@$N$, M, R, C and S are short for BLEU@$N$, METEOR, ROUGE-L, CIDEr-D and SPICE scores. All values are reported as percentage (\%). ${\Sigma}$ indicates an ensemble or fusion.}
  \resizebox{0.98\textwidth}{!}{
  \begin{tabular}{l | c c c c c c | c c c c c c}
      \toprule
      % \hline
      Model & \multicolumn{6}{c|}{\textbf{Cross-Entropy Loss}} & \multicolumn{6}{c}{\textbf{CIDEr-D Score Optimization}} \\
      \hline
      Metric         & B@1    & B@4    & M       & R      & C       & S      & B@1    & B@4    & M       & R      & C       & S    \\
  \midrule
  % \hline
   &\multicolumn{12}{c}{\textbf{Single Model}} \\
   \midrule
  % \hline
   LSTM \cite{Vinyals2015Show}       & ~-~      & ~29.6~   & ~25.2~    & ~52.6~   & ~94.0~    & ~-~      & ~-~      & ~31.9~   & ~25.5~    & ~54.3~   & ~106.3~   & -    \\
  SCST \cite{rennie2017self-critical}   & -      & 30.0   & 25.9    & 53.4   & 99.4    & -      & -      & 34.2   & 26.7    & 55.7   & 114.0   & -    \\
  LSTM-A \cite{yao2017boosting}       & 75.4   & 35.2   & 26.9    & 55.8   & 108.8   & 20.0   & 78.6   & 35.5   & 27.3    & 56.8   & 118.3   & 20.8 \\
  Up-Down \cite{Anderson2018Bottom}      & 77.2   & 36.2   & 27.0    & 56.4   & 113.5   & 20.3   & 79.8   & 36.3   & 27.7    & 56.9   & 120.1   & 21.4 \\
  RFNet~\cite{jiang2018recurrent}&76.4 &35.8  &27.4 &56.8  &112.5  & 20.5 &79.1 &36.5  &27.7 &57.3  &121.9  & 21.2\\
  GCN-LSTM~\cite{yao2018exploring} & 77.3   & 36.8   & 27.9    & 57.0   & 116.3   & 20.9   & 80.5   & 38.2   & 28.5    & 58.3   & 127.6   & 22.0 \\
  SGAE~\cite{yang2019auto} & -  & -   & -    & -   & -   & -   & \textbf{80.8}  & 38.4   & 28.4    & 58.6   & 127.8   & 22.1 \\
  \hline
  AoANet (Ours)  &\textbf{77.4}&\textbf{37.2}&\textbf{28.4}&\textbf{57.5}&\textbf{119.8}&\textbf{21.3} &80.2&\textbf{38.9}&\textbf{29.2}&\textbf{58.8}&\textbf{129.8}&\textbf{22.4}\\
  \midrule
  
   &\multicolumn{12}{c}{\textbf{Ensemble/Fusion}} \\
  \midrule
  SCST~\cite{rennie2017self-critical}$^{\Sigma}$   & -      & 32.8   & 26.7    & 55.1   & 106.5    & -      & -      & 35.4   & 27.1    & 56.6   & 117.5   & -    \\
  RFNet~\cite{jiang2018recurrent}$^{\Sigma}$&77.4 &37.0  &27.9 &57.3  &116.3  & 20.8 &80.4 &37.9 &28.3 &58.3& 125.7 &21.7\\
  GCN-LSTM~\cite{yao2018exploring}$^{\Sigma}$       & 77.4   & 37.1   & 28.1    & 57.2~   & 117.1  & 21.1   & 80.9   & 38.3  & 28.6    & 58.5   & 128.7  & 22.1 \\
  SGAE~\cite{yang2019auto}$^{\Sigma}$ & -  & -   & -    & -   & -   & -   & 81.0  & 39.0  & 28.4    & 58.9   & 129.1   & 22.2 \\
  \hline
  AoANet (Ours)$^{\Sigma}$  &\textbf{78.7}&\textbf{38.1}&\textbf{28.5}&\textbf{58.2}&\textbf{122.7}&\textbf{21.7} &\textbf{81.6}&\textbf{40.2}&\textbf{29.3}&\textbf{59.4}&\textbf{132.0}&\textbf{22.8}\\
  \bottomrule
  \end{tabular}}
  \label{tab:COCO}
  \end{table*}

\begin{table*}[h]
  \centering
  \caption{Leaderboard of various methods on the online MS-COCO test server.}
  \label{tab:COCO_online}
  % \scalebox{0.86}{
    \resizebox{0.98\textwidth}{!}{
  \begin{tabular}{l| c c c c c c c c c c c c c c}
          \toprule
              Model & \multicolumn{2}{c}{BLEU-1}  & \multicolumn{2}{c}{BLEU-2}  & \multicolumn{2}{c}{BLEU-3} & \multicolumn{2}{c}{BLEU-4} &\multicolumn{2}{c}{METEOR} &\multicolumn{2}{c}{ROUGE-L} & \multicolumn{2}{c}{CIDEr-D}\\ \hline 
              Metric  &  c5 & c40 & c5 & c40 &  c5 & c40 & c5 & c40 & c5 & c40 & c5 & c40 & c5 & c40 \\ \hline
              SCST~\cite{rennie2017self-critical}      & 78.1 & 93.7 & 61.9 & 86.0 & 47.0 & 75.9 & 35.2 & 64.5 & 27.0 & 35.5 & 56.3 & 70.7 & 114.7 & 116.0  \\
              LSTM-A~\cite{yao2017boosting}   & 78.7 & 93.7 & 62.7 & 86.7 & 47.6 & 76.5 & 35.6 & 65.2 & 27.0 & 35.4 & 56.4 & 70.5 & 116.0 & 118.0  \\ 
              Up-Down~\cite{Anderson2018Bottom}    & 80.2 & 95.2 & 64.1 & 88.8 & 49.1 & 79.4 & 36.9 & 68.5 & 27.6 & 36.7 & 57.1 & 72.4 & 117.9& 120.5  \\ 
              RFNet~\cite{jiang2018recurrent}           & 80.4 & 95.0 & 64.9 & 89.3 & 50.1 & 80.1 & 38.0 & 69.2 & 28.2 & 37.2 & 58.2 & 73.1 & 122.9& 125.1    \\
              GCN-LSTM~\cite{yao2018exploring}           & - & - & 65.5 & 89.3 & 50.8 & 80.3 & 38.7 & 69.7 & 28.5 & 37.6 & 58.5 & 73.4 & 125.3& 126.5    \\
              SGAE~\cite{yang2019auto}    & \textbf{81.0} & \textbf{95.3}  & 65.6 & 89.5 & 50.7  & 80.4 & 38.5 & 69.7& 28.2 & 37.2 & 58.6 & 73.6 & 123.8 & 126.5\\ 
  \hline
              AoANet (Ours)    & \textbf{81.0} & 95.0  & \textbf{65.8} & \textbf{89.6} & \textbf{51.4}  & \textbf{81.3} & \textbf{39.4} & \textbf{71.2}& \textbf{29.1} & \textbf{38.5} & \textbf{58.9} & \textbf{74.5} & \textbf{126.9} & \textbf{129.6}\\ 
              \bottomrule
  \end{tabular}}
\end{table*}

The context vector $\boldsymbol{c}_t$ saves the decoding state and the newly acquired information, which is generated with the attended feature vector $\hat{\boldsymbol{a}}_t$ and the output $\boldsymbol{h}_t$ of an LSTM, where $\hat{\boldsymbol{a}}_t$ is the attended result from an attention module which could have a single head or multiple heads.

The LSTM in the decoder models the caption decoding process. Its input consists of the embedding of the input word at current time step, and a visual vector $(\bar{\boldsymbol{a}}+\boldsymbol{c}_{t-1})$, where $\bar{\boldsymbol{a}} = \frac{1}{k}\sum_i \boldsymbol{a}_i$ denotes the mean pooling of $\A$ and $\boldsymbol{c}_{t-1}$ denotes the context vector at previous time step ($\boldsymbol{c}_{-1}$ is initialized to zeros at the beginning step):
\begin{align}
  \boldsymbol{x}_t &= [ W_{e}\Pi_t,\bar{\boldsymbol{a}}+\boldsymbol{c}_{t-1}] \\
  \boldsymbol{h}_t, \boldsymbol{m}_t &= \textrm{LSTM}(\boldsymbol{x}_t, \boldsymbol{h}_{t-1}, \boldsymbol{m}_{t-1})
\end{align}
where $W_{e} \in \mathbb{R}^{E \times \vert\Sigma\vert}$ is a word embedding matrix for a vocabulary $\Sigma$, and $\Pi_t $ is one-hot encoding of the input word $w_t$ at time step $t$.

As shown in Figure \ref{fig:decoder}, for the AoA decoder, $\boldsymbol{c}_t$ is obtained from an AoA module, notated as $\textrm{AoA}^D$:
\begin{align}
  \boldsymbol{c}_t =  \textrm{AoA}^D( f_{mh-att}, W^{Q_d}[\boldsymbol{h}_t], W^{K_d} A, W^{V_d} A )
\end{align}
where $W^{Q_e}, W^{K_e}, W^{V_e} \in \mathbb{R}^{D \times D}$; $\boldsymbol{h}_t,\boldsymbol{m}_t \in \mathbb{R}^D$ the hidden states of the LSTM and $\boldsymbol{h}_t$ serves as the attention query.

\newcolumntype{P}[1]{>{\centering\arraybackslash}p{#1}}
\newcolumntype{M}[1]{>{\centering\arraybackslash}m{#1}}
\begin{table}[htbp]
  \centering
  \caption{Examples of captions generated by AoANet and a baseline model as well as the corresponding ground truths.}
  \resizebox{0.46\textwidth}{!}{
   \begin{tabular}{  m{2.6cm} | m{5.0cm}  }
 \toprule
     Image & Captions \\
     \hline 
     \begin{minipage}{.15\textwidth}
       \includegraphics[width=\linewidth]{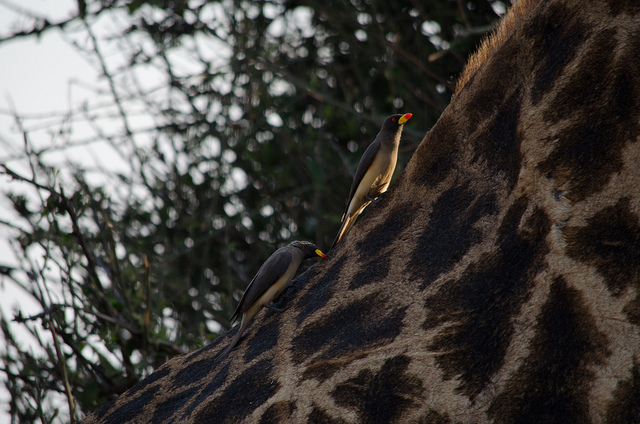}
     \end{minipage} 
     & 
     \begin{minipage}[h]{5.0cm}\scriptsize
         \textbf{AoANet}: Two birds sitting on top of a giraffe.\\
         \textbf{Baseline}: A bird sitting on top of a tree.\\
         \textbf{GT}1. Two birds going up the back of a giraffe. \\
         \textbf{GT}2. A large giraffe that is walking by some trees. \\
         \textbf{GT}3. Two birds are sitting on a wall near the bushes. \\
     \end{minipage} \\
     
     \hline
     \begin{minipage}{.15\textwidth}
       \includegraphics[width=\linewidth]{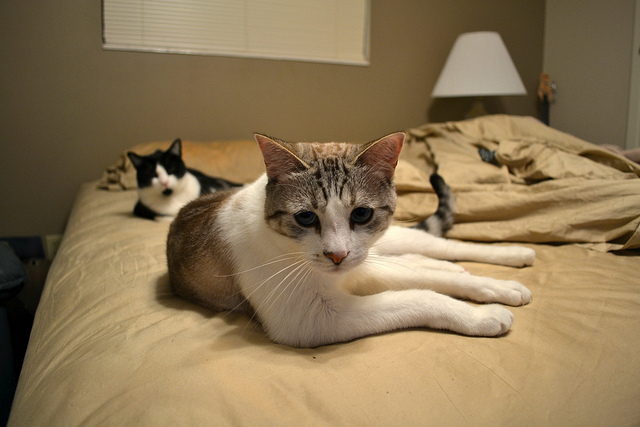}
     \end{minipage} 
     & 
     \begin{minipage}[h]{5.0cm}\scriptsize
         \textbf{AoANet}: Two cats laying on top of a bed. \\
         \textbf{Baseline}: A black and white cat laying on top of a bed.\\
         \textbf{GT}1. A couple of cats laying on top of a bed. \\
         \textbf{GT}2. Two cats laying on a big bed and looking at the camera. \\
         \textbf{GT}3. A couple of cats on a mattress laying down. \\
        %  \textbf{GT}4. A couple of cats lay down on a bed. \\
        %  \textbf{GT}5. Two cats who are laying on a bed. \\ 
     \end{minipage}\\

     \hline
     \begin{minipage}{.15\textwidth}
       \includegraphics[width=\linewidth]{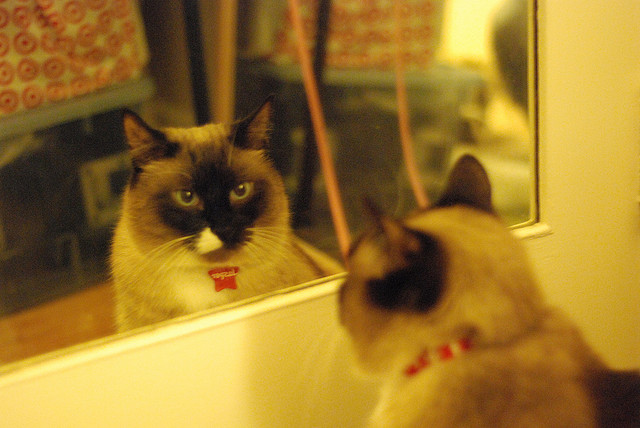}
     \end{minipage} 
     & 
     \begin{minipage}[h]{5.0cm}\scriptsize
         \textbf{AoANet}: A cat looking at its reflection in a mirror. \\
         \textbf{Baseline}: A cat is looking out of a window.\\
         \textbf{GT}1. A cat looking at his reflection in the mirror. \\
         \textbf{GT}2. A cat that is looking in a mirror. \\
         \textbf{GT}3. A cat looking at itself in a mirror. \\
        %  \textbf{GT}4. A cat looking at itself adoringly in a mirror. \\
        %  \textbf{GT}5. A cat stares at itself in a mirror. \\ 
     \end{minipage}\\
     
     \hline
     \begin{minipage}{.15\textwidth}
       \includegraphics[width=\linewidth]{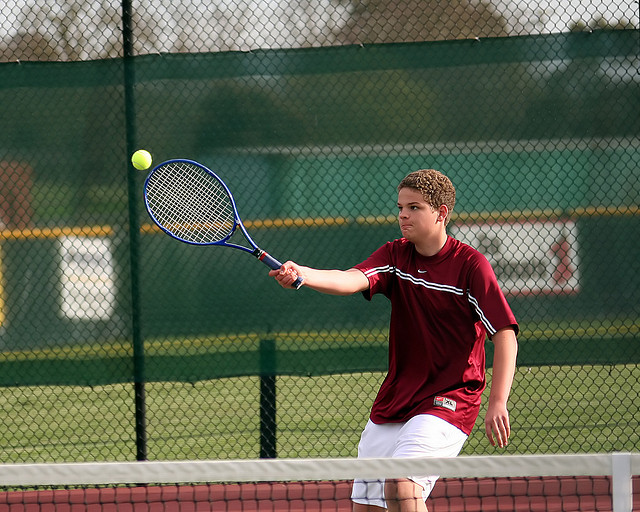}
     \end{minipage} 
     & 
     \begin{minipage}[h]{5.0cm}\scriptsize
         \textbf{AoANet}: A young boy hitting a tennis ball with a tennis racket. \\
         \textbf{Baseline}: A young man holding a tennis ball on a court.\\
         \textbf{GT}1. A guy in a maroon shirt is holding a tennis racket out to hit a tennis ball. \\
         \textbf{GT}2. A man on a tennis court that has a racquet. \\
         \textbf{GT}3. A boy hitting a tennis ball on the tennis court. \\
        %  \textbf{GT}4. A person hitting a tennis ball with a tennis racket. \\
        %  \textbf{GT}5. A boy attempts to hit the tennis ball\\ with the racquet. \\ 
     \end{minipage} \\

 \bottomrule
   \end{tabular}}
   \label{caption-example}
 \end{table}
\subsection{Training and Objectives}
\noindent\textbf{Training with Cross Entropy Loss.}
We first train AoANet by optimizing the cross entropy (XE) loss $L_{XE}$:
\begin{align}
  L_{XE}(\theta) &= -\sum_{t=1}^{T}\log(p_{\theta}(\y_t^* \mid \y_{1:t-1}^*))
\end{align}
where $\y_{1:T}^*$ denotes the target ground truth sequence.

\noindent\textbf{CIDEr-D Score Optimization.}
Then we directly optimize the non-differentiable metrics with Self-Critical Sequence Training~\cite{rennie2017self-critical} (SCST):
\begin{align}
  \label{eqn:ciderloss}
  L_{RL}(\theta) &= -{\ensuremath{\textbf{E}_{\y_{1:T} \sim p_\theta}\!\left[{r({\y}_{1:T})}\right]}}
  \end{align}
  \noindent where the reward $r(\cdot)$ uses the score of some metric (\eg CIDEr-D~\cite{Vedantam2015CIDEr}).
  The gradients can be approximated:

\begin{align}
  \nabla_{\theta}L_{RL}(\theta) \approx -(r({\y^s}_{1:T})-r({\hat{\y}}_{1:T}))     \nabla_{\theta}\log p_\theta(\y_{1:T}^s)
\end{align}

\noindent $\y^s$ means it's a result sampled from probability distribution, while $\hat{\y}$ indicates a result of greedy decoding.

\subsection{Implementation Details}
We employ a pre-trained Faster-RCNN~\cite{Ren2015Faster} model on ImageNet~\cite{deng2009imagenet} and Visual Genome~\cite{krishna2017visual} to extract bottom-up feature vectors of images \cite{Anderson2018Bottom}. The dimension of the original vectors is 2048 and we project them to a new space with the dimension of $D=1024$, which is also the hidden size of the LSTM in the decoder.
As for the training process, we train AoANet under XE loss for 30 epochs with a mini batch size of 10, and ADAM~\cite{Kingma2014Adam} optimizer is used with a learning rate initialized by 2e-4 and annealed by 0.8 every 3 epochs. We increase the scheduled sampling probability by 0.05 every 5 epochs \cite{Bengio2015Scheduled}. We optimize the CIDEr-D score with SCST for another 15 epochs with an initial learning rate of 2e-5 and annealed by 0.5 when the score on the validation split does not improve for some training steps.

\section{Experiments}
\subsection{Dataset}
We evaluate our proposed method on the popular MS COCO dataset \cite{Lin2014Microsoft}.
MS COCO dataset contains 123,287 images labeled with 5 captions for each, including 82,783 training images and 40,504 validation images. MS COCO provides 40,775 images as test set for online evaluation as well.
The offline ``Karpathy'' data split \cite{Karpathy2015Deep} is used for the offline performance comparisons, where 5,000 images are used for validation, 5,000 images for testing and the rest for training.
We convert all sentences to lower case, and drop the words that occur less than 5 times and end up with a vocabulary of 10,369 words.
% TODO url coco-caption
We use different metrics, including BLEU~\cite{Papineni2002A}, METEOR~\cite{Satanjeev2005METEOR}, ROUGE-L~\cite{Flick2004ROUGE}, CIDEr-D~\cite{Vedantam2015CIDEr} and SPICE~\cite{Anderson2016SPICE}, to evaluate the proposed method and compare with other methods.
All the metrics are computed with the publicly released code\footnote{https://github.com/tylin/coco-caption}.

\subsection{Quantitative Analysis}
\textbf{Offline Evaluation.} We report the performance on the offline test split of our model as well as the compared models in Table~\ref{tab:COCO}.
The models include: LSTM~\cite{Vinyals2015Show}, which encodes the image using CNN and decodes it using LSTM; SCST~\cite{rennie2017self-critical}, which employs a modified visual attention and is the first to use SCST to directly optimize the evaluation metrics; Up-Down~\cite{Anderson2018Bottom}, which employs a two-LSTM layer model with bottom-up features extracted from Faster-RCNN;
RFNet~\cite{jiang2018recurrent}, which fuses encoded features from multiple CNN networks;
GCN-LSTM~\cite{yao2018exploring}, which predicts visual relationships between every two entities in the image and encodes the relationship information into feature vectors;
and SGAE~\cite{yang2019auto}, which introduces auto-encoding scene graphs into its model.

For fair comparison, all the models are first trained under XE loss and then optimized for CIDEr-D score. For the XE loss training stage in Table~\ref{tab:COCO}, it can be seen that our single model achieves the highest scores among all compared methods in terms of all metrics even comparing with the ensemble of their models.
As for the CIDEr-D score optimization stage, an ensemble of 4 models with different parameter initialization of AoANet outperforms all other models and sets a new state-of-the-art performance of 132.0 CIDEr-D score.

\textbf{Online Evaluation.} We also evaluate our model on the online COCO test server\footnote{https://competitions.codalab.org/competitions/3221\#results} in Table~\ref{tab:COCO_online}. The results of AoANet are evaluated by an ensemble of 4 models trained on the ``Karpathy'' training split. AoANet achieves the highest scores for most metrics except a slightly lower one for BLEU-1 (C40).

\subsection{Qualitative Analysis}

Table \ref{caption-example} shows a few examples with images and captions generated by our AoANet and a strong baseline as well as the human-annotated ground truths. We derive the baseline model by
re-implementing the Up-Down~\cite{Anderson2018Bottom} model with the settings of AoANet.
From these examples, we find that the baseline model generates captions which are in line with the logic of language but inaccurate for the image content, while AoANet 
generates accurate captions in high quality. More specifically, our AoANet is superior in the following two aspects: 1) AoANet counts objects of the same kind more accurately. There are two birds/cats in the image of the first/second example. However, the baseline model finds only one while our AoANet counts correctly; 2) AoANet figures out the interactions of objects in an image. For example, AoANet knows that the birds are on top of a \emph{giraffe} but not the \emph{tree}, in the first example; the boy is \emph{hitting} the tennis ball with a racket but not \emph{holding}, in the fourth example. AoANet has these advantages because it can figure out the connections among objects and also knows how they are connected: in the encoder, the refining module uses self-attention to seek interactions among objects and uses AoA to measure how well they are related;
in the decoder, AoA helps to filter out irrelative objects which don't have the required interactions and only keeps the related ones.
While the baseline model generates captions which are logically right but might not match the image contents.

\begin{figure}[t] 
          \subfigure[Refining - without AoA]{  
            \begin{minipage}[b]{0.54\linewidth}
            \centering                                       
            \includegraphics[scale=0.45]{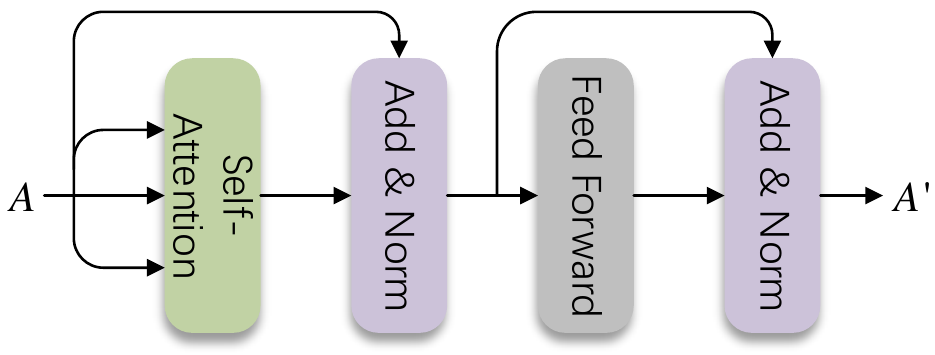}   
            \label{base-encoder}    
            \end{minipage}
          }
          \subfigure[Refining - with AoA]{  
            \begin{minipage}[b]{0.4\linewidth}
            \centering                                       
            \includegraphics[scale=0.42]{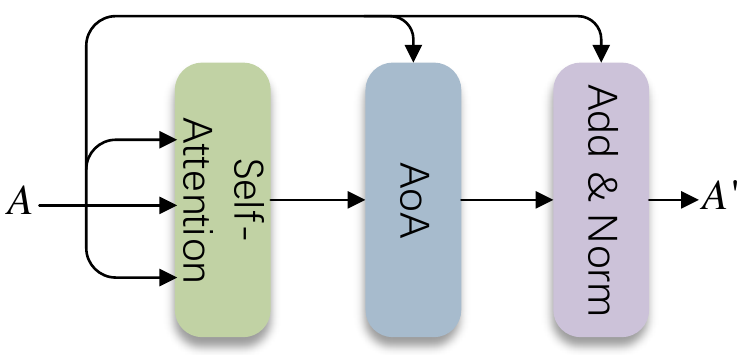}   
            \label{aoa-encoder}    
            \end{minipage}
          }     
    \caption{Refining modules w/o and w/ AoA.} 
    \label{fig:encoder:ablation}         
\end{figure}

\begin{figure}[t]            
    \subfigure[Base]{  
      \begin{minipage}[b]{0.3\linewidth}
      \centering                                       
      \includegraphics[scale=0.6]{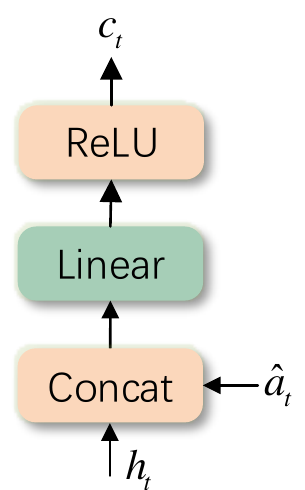}   
      \label{base-decoder}   
      \end{minipage}
    }
    \subfigure[LSTM]{  
      \begin{minipage}[b]{0.3\linewidth}
      \centering                                       
      \includegraphics[scale=0.6]{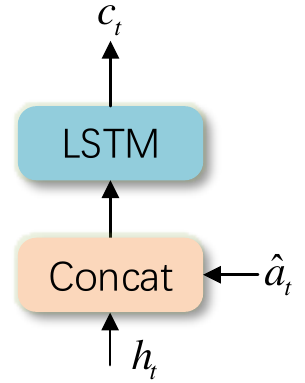}   
      \label{lstm-decoder}   
      \end{minipage}
    }              
    \subfigure[AoA]{  
      \begin{minipage}[b]{0.3\linewidth}
      \centering                                       
      \includegraphics[scale=0.6]{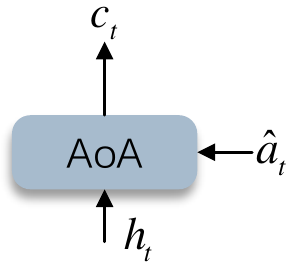}   
      \label{aoa-decoder}  
      \end{minipage}
    }            

    \caption{Different schemes for decoders to model $\c_t$.} 
    \label{fig:encoder:ablation}      
\end{figure}

\begin{figure*}[t] 
  \subfigure[Base – A teddy bear sitting \textcolor{red}{on a book on a book}.]{  
    \begin{minipage}{\linewidth}
    \centering                                       
    \includegraphics[scale=0.35]{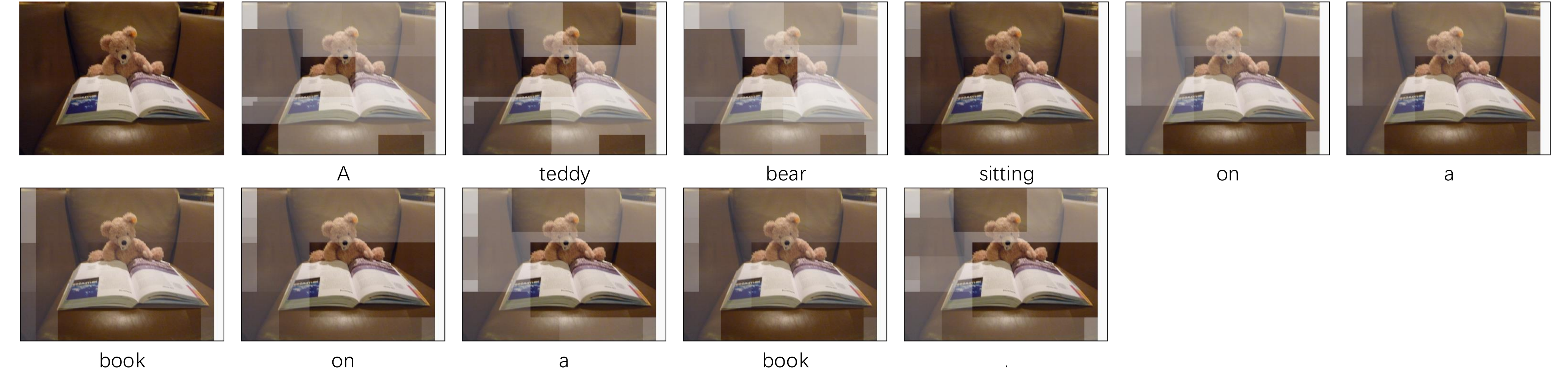}   
    \label{vis-base} 
    \end{minipage}
  }
  \subfigure[AoA – A teddy bear sitting \textcolor{red}{on a chair with a book}.]{  
    \begin{minipage}{\linewidth}
    \centering                                       
    \includegraphics[scale=0.35]{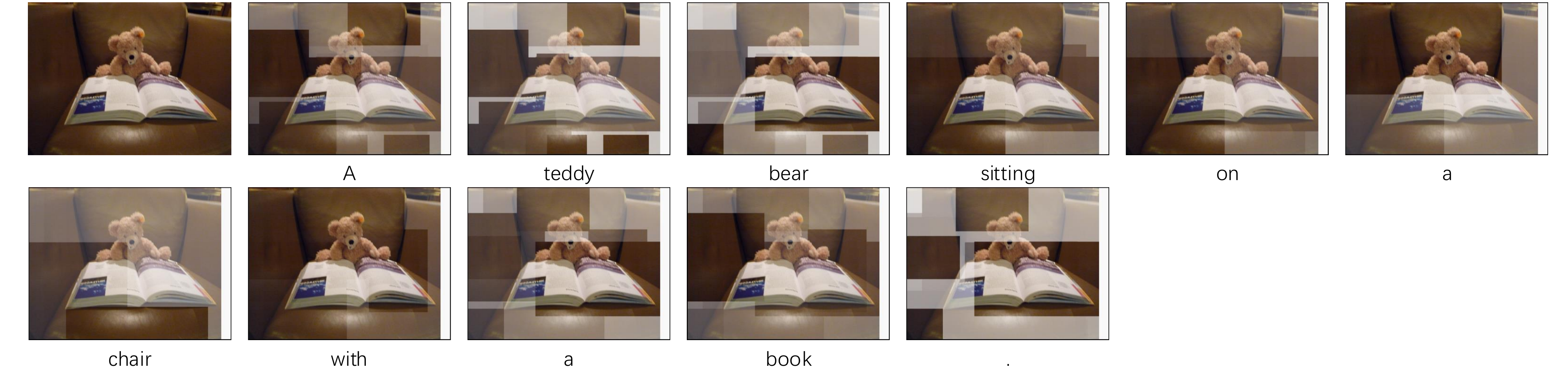}   
    \label{vis-aoa} 
    \end{minipage}
  }     
\caption{Visualization of attention regions in the caption generation process for the ``base'' model and ``decoder with AoA''. The ``base'' model can be easily misled by irrelevant attention while ``decoder with AoA'' is less likely so.} 
\label{fig:vis}         
\end{figure*}

\begin{table}[t]
  \centering
  \caption{Settings and results of ablation studies. The results are reported after XE training stage.}
  \resizebox{0.48\textwidth}{!}{
  \begin{tabular}{l c c c c c c}
  \toprule
    Model & B@1 & B@4 & R & C\\
    \hline
     Base  & 75.7  &34.9&56.0&109.5 \\
    
    \hline
    + Enc: Refine (w/o AoA)   & \textbf{77.0}  &35.6&56.4&112.5 \\
    + Enc: Refine (w/ AoA)   &76.7 &\textbf{36.1}&\textbf{56.7}&\textbf{114.5} \\
    \midrule
    + Dec: LSTM  & 76.8&35.9&56.6&113.5 \\
    + Dec: AoA  & 76.6  &35.8&56.6&113.8 \\
    + Dec: LSTM + AoA  &  \multicolumn{4}{c}{\emph{unstable training process}} \\
    + Dec: MH-Att  & 75.8&34.8&56.0&109.6 \\
    + Dec: MH-Att, LSTM  & 76.6 &35.8&\textbf{56.7}&113.8 \\
    + Dec: MH-Att, AoA   &\textbf{76.9}&\textbf{36.1}&56.6&\textbf{114.3} \\
    \hline
    Full: AoANet &\textbf{77.4}&\textbf{37.2}&\textbf{57.5}&\textbf{119.8} \\
  \bottomrule
  \end{tabular}}
  \label{ablative-table}
\end{table}

\subsection{Ablative Analysis}
To quantify the impact of the proposed AoA module, we compare AoANet against a set of other ablated models with various settings.
We first design the ``base'' model which doesn't have a refining module in its encoder and adopts a ``base'' decoder in Figure~\ref{base-decoder}, using a linear transformation to generate the context vector $\c_t$.

\textbf{Effect of AoA on the encoder.} To evaluate the effect of applying AoA to the encoder,
% we design a refining module with solely self-attention and a following feed-forward transition, in Figure~\ref{base-encoder}.
we design a refining module without AoA, which contains a self-attention module and a following feed-forward transition, in Figure~\ref{base-encoder}.
From Table~\ref{ablative-table} we observe that refining the feature representations brings positive effects, and adding a refining module without AoA improves the CIDEr-D score of ``base'' by 3.0.
We then apply AoA to the attention mechanism in the refining module and we drop the feed-forward layer. The results show that our AoA further improves the CIDEr-D score by 2.0.

\textbf{Effect of AoA on the decoder.} We compare the performance of using different schemes to model the context vector $\c_t$: ``base'' (Figure~\ref{base-decoder}), via a linear transformation; ``LSTM'' (Figure~\ref{lstm-decoder}), via an LSTM; AoA (Figure~\ref{aoa-decoder}), by applying AoA. We conduct experiments with both single attention and multi-head attention (MH-Att). From Table~\ref{ablative-table}, we observe that replacing single attention with multi-head attention brings slightly better performances. Using LSTM improves the performance of the base model, and AoA further outperforms LSTM. Compared to LSTM, which uses some memories (hidden states) and gates to model attention results in a sequential manner, AoA is more light-weighted as it involves only two linear transformation and requires little computation. Even so, AoA still outperforms LSTM. We also find that the training process of ``LSTM + AoA'' (building AoA upon LSTM) is unstable and could reach a sub-optimal point, which indicates that stacking more gates doesn't provide further performance improvements.

To qualitatively show the effect of AoA, we visualize the caption generation process in Figure~\ref{fig:vis} with attended image regions for each decoding time step. Two models are compared: the ``base'' model, which doesn't incorporate the AoA module, and ``decoder with AoA'', which employs an AoA module in its caption decoder. Observing the attended image regions in Figure~\ref{fig:vis}, we find that the attention module isn't always reliable for the caption decoder to generate a word, and directly using the attention result might result in wrong captions. In the example, the book is attended by the base model when generating the caption fragment \emph{``A teddy bear sitting on a ...''}. As a result, the base model outputs \emph{``book''} for the next word, which is not consistent to what the image shows since the teddy bear is actually sitting on a \emph{chair} but not on a \emph{book}.
In contrast, ``decoder with AoA'' is less likely to be misled by irrelevant attention results, because the AoA module in it adds another attention on the attention result, which suppress the irrelevant/misleading information and keeps only the useful.

\subsection{Human Evaluation} We follow the practice in \cite{yang2019auto} and invited 30 evaluators to evaluate 100 randomly selected images. For each image, we show the evaluators two captions generated by ``decoder with AoA'' and the ``base'' model in random order, and ask them which one is more descriptive. The percentages of ``decoder with AoA'', ``base'', and \emph{comparative} are 49.15\%, 21.2\%, and 29.65\% respectively, 
 which shows the effectiveness of AoA as confirmed by the evaluators.

\subsection{Generalization}
To show the general applicability of AoA, we perform experiments on a video captioning dataset, MSR-VTT~\cite{Xu_2016_CVPR}:
we use ResNet-101 to extract feature vectors from sampled 20 frames of each video and then pass them to a bi-LSTM and a decoder, ``base'' or ``decoder with AoA''.
We find that ``decoder with AoA'' improves ``base'' from BLEU-4: 33.53,CIDEr-D: 38.83, ROUGE-L 56.90 to 37.22, 42.44, 58.32, respectively, which shows that AoA is also promising for other tasks which involve attention mechanisms.

\section{Conclusion}
In this paper,
we propose the \emph{Attention on Attention} (AoA) module, an extension to conventional attention mechanisms, to address the irrelevant attention issue.
Furthermore, we propose AoANet for image captioning by applying AoA to both the encoder and decoder. More remarkably, we achieve a new state-of-the-art performance with AoANet.
Extensive experiments conducted on the MS COCO dataset demonstrate the superiority and general applicability of our proposed AoA module and AoANet.

\section*{Acknowledgment}
This project was supported by Shenzhen Key Laboratory for Intelligent Multimedia and Virtual Reality (ZDSYS201703031405467), National Natural Science Foundation of China (NSFC, No.U1613209, 61872256, 61972217), and National Engineering Laboratory for Video Technology - Shenzhen Division.
We would also like to thank Qian Wu, Yaxian Xia and Qixiang Ye, as well as the anonymous reviewers for their insightful comments.

\bibliographystyle{ieee_fullname}
\bibliography{5657}

\clearpage
\appendix



\section*{Supplementary material}

| Image | Generated captions | Ground truths |
|---|---|---|
| 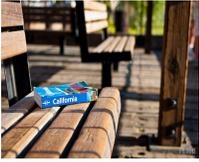 | **Base model**: A wooden bench sitting on top of a wooden bench.<br>**Encoder (w/ refining, w/o AoA)**: A box of books sitting on top of a wooden bench.<br>**Encoder (w/ refining, w/ AoA)**: A book sitting on top of a wooden bench.<br>**LSTM decoder (H=1)**: A wooden bench sitting next to a wooden bench.<br>**AoA decoder (H=1)**: A wooden bench sitting on top of a wooden bench.<br>**Base decoder (H=8)**: A row of books sitting on a wooden bench.<br>**LSTM decoder (H=8)**: A wooden bench sitting on the side of a wooden bench.<br>**AoA decoder (H=8)**: A row of benches sitting on a wooden bench.<br>**Full (AoANet)**: A book sitting on top of a wooden bench. | GT1: A book on a wooden bench on a street.<br>GT2: A book rest on one of the benches.<br>GT3: A california travel guide book on a park bench.<br>GT4: A big entitled california is on the bench.<br>GT5: A book sitting on top of a wooden desk. |
| 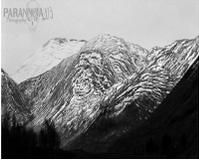 | **Base model**: A mountain with a mountain in the background.<br>**Encoder (w/ refining, w/o AoA)**: A view of a mountain with a mountain in the background.<br>**Encoder (w/ refining, w/ AoA)**: A black and white photo of a mountain.<br>**LSTM decoder (H=1)**: A mountain in the background with a mountain.<br>**AoA decoder (H=1)**: A black and white photo of a mountain on a mountain.<br>**Base decoder (H=8)**: A view of a mountain with a mountain.<br>**LSTM decoder (H=8)**: A mountain with a mountain in the background.<br>**AoA decoder (H=8)**: A view of a mountain with a mountain in the background.<br>**Full (AoANet)**: A black and white photo of a mountain covered with snow. | GT1: A black and white picture of hills or mountains.<br>GT2: A black and white photo of mountains and the sky.<br>GT3: A black and white photograph of mountains with snow.<br>GT4: A black and white picture of the top view of a snow capped mountain somewhere.<br>GT5: A large mountain range partially covered with snow. |
| 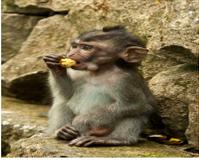 | **Base model**: A brown bear eating a piece of food.<br>**Encoder (w/ refining, w/o AoA)**: A brown bear eating a piece of food.<br>**Encoder (w/ refining, w/ AoA)**: A monkey eating a doughnut in a rock.<br>**LSTM decoder (H=1)**: A brown bear sitting on top of a rock eating a piece of food.<br>**AoA decoder (H=1)**: A brown bear sitting on top of a rock eating a snack.<br>**Base decoder (H=8)**: A brown bear eating a banana in its mouth.<br>**LSTM decoder (H=8)**: A brown bear eating a piece of food.<br>**AoA decoder (H=8)**: A brown bear eating a piece of food on a rock.<br>**Full (AoANet)**: A monkey sitting on rocks eating a piece of food. | GT1: Small monkey eating fruit sitting on a rock.<br>GT2: A small monkey sitting and eating in its habitat.<br>GT3: A little monkey sitting next to a rock eating a piece of fruit.<br>GT4: A monkey is eating while sitting on a rock.<br>GT5: A monkey sits downs while eating some food. |
| 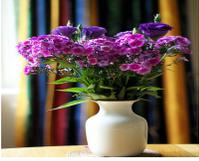 | **Base model**: A vase with purple flowers sitting on a table.<br>**Encoder (w/ refining, w/o AoA)**: A vase with flowers sitting on top of a table.<br>**Encoder (w/ refining, w/ AoA)**: A white vase with purple flowers sitting on a table.<br>**LSTM decoder (H=1)**: A vase with flowers sitting on top of a table.<br>**AoA decoder (H=1)**: A vase filled with purple flowers sitting on a table.<br>**Base decoder (H=8)**: A white vase with purple flowers sitting on a table.<br>**LSTM decoder (H=8)**: A vase with flowers sitting on a table.<br>**AoA decoder (H=8)**: A white vase filled with purple flowers on a table.<br>**Full (AoANet)**: A white vase filled with purple flowers on a table. | GT1: A vase of light and dark purple flowers in a white vase.<br>GT2: A close up of a vase with many flowers.<br>GT3: A bouquet of fresh flowers in a cream colored vase.<br>GT4: A small white vase of purple flowers on a table.<br>GT5: A white vase filled with purple flowers on top of a table. |
| 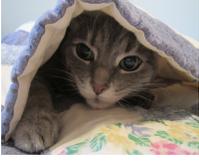 | **Base model**: A cat laying on top of a bed.<br>**Encoder (w/ refining, w/o AoA)**: A black and white cat laying on a bed.<br>**Encoder (w/ refining, w/ AoA)**: A cat laying on top of a blanket.<br>**LSTM decoder (H=1)**: A cat laying on top of a bed.<br>**AoA decoder (H=1)**: A cat laying on top of a blanket.<br>**Base decoder (H=8)**: A cat laying on top of a bed.<br>**LSTM decoder (H=8)**: A cat laying on top of a bed.<br>**AoA decoder (H=8)**: A cat laying on top of a blanket.<br>**Full (AoANet)**: A cat laying on a bed with a blanket. | GT1: A cat under a blanket looking at something.<br>GT2: A small cat peeking its head from under a blanket.<br>GT3: A cute kitten peeking out from under a quilt.<br>GT4: A gray tiger cat sleeping on a bed under a blanket.<br>GT5: A cat is peeping out from under a blanket. |
| 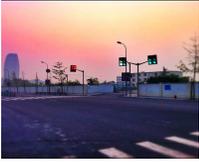 | **Base model**: A traffic light on the side of a street.<br>**Encoder (w/ refining, w/o AoA)**: A traffic light on the side of a street.<br>**Encoder (w/ refining, w/ AoA)**: A group of traffic lights on a city street.<br>**LSTM decoder (H=1)**: A traffic light on a city street with a building.<br>**AoA decoder (H=1)**: A traffic lights on a city street with buildings.<br>**Base decoder (H=8)**: A traffic light on the side of a city street.<br>**LSTM decoder (H=8)**: A traffic light on the side of a street.<br>**AoA decoder (H=8)**: A traffic light on a city street with a city.<br>**Full (AoANet)**: A street with traffic lights and buildings in the background. | GT1: A couple of traffic lights hanging over a city street.<br>GT2: Traffic lights shine over an empty intersection at twilight.<br>GT3: A street with traffic lights, wall and buildings.<br>GT4: The electronic traffic signals are lit up during dawn.<br>GT5: Empty city intersection in the early morning hours. |
| 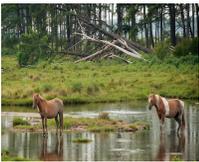 | **Base model**: A brown horse standing in the water next to a lake.<br>**Encoder (w/ refining, w/o AoA)**: Two horses are standing in the water.<br>**Encoder (w/ refining, w/ AoA)**: Two horses are standing in the water.<br>**LSTM decoder (H=1)**: A horse standing in the water in a field.<br>**AoA decoder (H=1)**: A horse standing in the water near a river.<br>**Base decoder (H=8)**: A brown horse standing in the water.<br>**LSTM decoder (H=8)**: A horse standing in the water in a field.<br>**AoA decoder (H=8)**: A horse standing in the water next to a river.<br>**Full (AoANet)**: Two horses walking in the water in a field. | GT1: A couple of horses standing in a river next to an island.<br>GT2: Two horses standing in a stream next to a wooded area.<br>GT3: Two horses are walking through a river together.<br>GT4: Horses standing in shallow water in a wooded area.<br>GT5: Two horses that are standing in some water. |

Figure 9: Captions generated by the ablated models and AoANet, as well as the corresponding ground truths.

| Image | Generated captions | Ground truths |
|---|---|---|
| 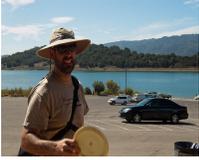 | **Base model**: A man standing in front of a car with a UNK.<br>**Encoder (w/ refining, w/o AoA)**: A man wearing a hat standing next to a beach with a frisbee.<br>**Encoder (w/ refining, w/ AoA)**: A man standing in a parking lot with a frisbee.<br>**LSTM decoder (H=1)**: A man is standing next to a car on a water.<br>**AoA decoder (H=1)**: A man is standing on a car with a frisbee.<br>**Base decoder (H=8)**: A man standing in a parking lot with a frisbee.<br>**LSTM decoder (H=8)**: A man sitting on a parking lot with a boat.<br>**AoA decoder (H=8)**: A man standing in a parking lot with a frisbee.<br>**Full (AoANet)**: A man wearing a hat holding a frisbee in a parking lot. | GT1: A man in holds a frisbee in the parking lot by the river.<br>GT2: The man found the frisbee by the parked cars.<br>GT3: A man holding a frisbee in a parking lot near water.<br>GT4: A man holding a white object in his hand.<br>GT5: A man wearing a hat and holding an object in his hand. |
| 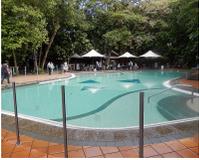 | **Base model**: A pool with a pool with people and umbrellas.<br>**Encoder (w/ refining, w/o AoA)**: A group of people standing in a park with umbrellas.<br>**Encoder (w/ refining, w/ AoA)**: A group of people standing in a pool with umbrellas.<br>**LSTM decoder (H=1)**: A pool with a pool with a pool with a pool.<br>**AoA decoder (H=1)**: A swimming pool with a pool with people.<br>**Base decoder (H=8)**: A pool with a pool with umbrellas on a pool.<br>**LSTM decoder (H=8)**: A swimming pool with a bunch of chairs and a table.<br>**AoA decoder (H=8)**: A swimming pool with tables and chairs with umbrellas.<br>**Full (AoANet)**: A swimming pool with umbrellas and people standing. | GT1: The clear pool stood empty as people stood around nearby.<br>GT2: Some people walking around a swimming pool and trees.<br>GT3: Large swimming pool with tents near the trees.<br>GT4: A large in ground swimming pool near tents.<br>GT5: A pool with several visitors sitting beside it. |
| 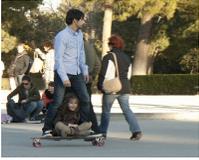 | **Base model**: A group of people walking down a street with skateboards.<br>**Encoder (w/ refining, w/o AoA)**: A man and a child riding on a skateboard.<br>**Encoder (w/ refining, w/ AoA)**: A man and a little girl on a skateboard.<br>**LSTM decoder (H=1)**: A group of people riding skateboards down a street.<br>**AoA decoder (H=1)**: A man and a child on a skateboard in the street.<br>**Base decoder (H=8)**: A group of people riding skateboards down a street.<br>**LSTM decoder (H=8)**: A man and a woman riding a skateboard down a street.<br>**AoA decoder (H=8)**: A man and a child on a skateboard.<br>**Full (AoANet)**: A man and a child sitting on a skateboard. | GT1: A girl hitching a ride with her dad on a skateboard.<br>GT2: A man riding a skateboard has a child between his legs.<br>GT3: A girl sits between a mans legs on a skateboard.<br>GT4: A man riding a skateboard next to a woman.<br>GT5: The skateboarder has a passenger on his skateboard. |
| 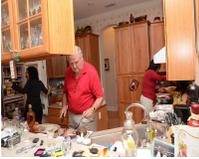 | **Base model**: A man and a woman standing in a kitchen preparing food.<br>**Encoder (w/ refining, w/o AoA)**: A group of people standing in a kitchen.<br>**Encoder (w/ refining, w/ AoA)**: A group of people standing in a kitchen preparing food.<br>**LSTM decoder (H=1)**: A man and a woman standing in a kitchen.<br>**AoA decoder (H=1)**: A group of people standing in a kitchen.<br>**Base decoder (H=8)**: A man and a woman standing in a kitchen.<br>**LSTM decoder (H=8)**: A man and a woman standing in a kitchen.<br>**AoA decoder (H=8)**: A man and a woman standing in a kitchen preparing food.<br>**Full (AoANet)**: A group of people standing in a kitchen preparing food. | GT1: A picture of an older man cutting a turkey on the counter.<br>GT2: A group of people that are standing in the kitchen.<br>GT3: A group of people standing over some food.<br>GT4: Three people agree in the kitchen near a cluttered bar.<br>GT5: A number of people in a kitchen preparing food. |
| 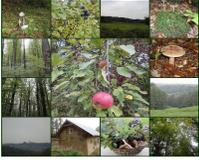 | **Base model**: A picture of a garden with a plant in a window.<br>**Encoder (w/ refining, w/o AoA)**: A bunch of apples in a window with a window.<br>**Encoder (w/ refining, w/ AoA)**: A garden with a bunch of apples and a window.<br>**LSTM decoder (H=1)**: A picture of a bunch of plants in a house.<br>**AoA decoder (H=1)**: A collage of pictures of a house with a tree.<br>**Base decoder (H=8)**: A group of plants in a window with a table.<br>**LSTM decoder (H=8)**: A picture of a tree with a bunch of apples.<br>**AoA decoder (H=8)**: A collage of pictures of a garden with a tree.<br>**Full (AoANet)**: A collage of pictures of a house with a flower. | GT1: Multiple images are seen showing forest scenes and a small cottage.<br>GT2: A collection of different plants and fields, with a tomato plant being the most prominent.<br>GT3: Various types of pictures that are of different plants.<br>GT4: Colorful collection of pictures of a wooded area.<br>GT5: A collage of several photos show outdoor scenes. |
| 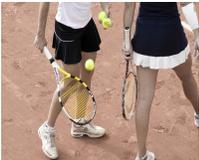 | **Base model**: A man hitting a tennis ball with a tennis racket.<br>**Encoder (w/ refining, w/o AoA)**: A group of people playing tennis on a tennis court.<br>**Encoder (w/ refining, w/ AoA)**: A group of people playing tennis on a tennis court.<br>**LSTM decoder (H=1)**: A group of people playing tennis on a tennis court.<br>**AoA decoder (H=1)**: A couple of women standing on a tennis court with tennis balls.<br>**Base decoder (H=8)**: A couple of people holding tennis rackets at a tennis ball.<br>**LSTM decoder (H=8)**: A woman hitting a tennis ball with a tennis racket.<br>**AoA decoder (H=8)**: Two women playing tennis on a tennis court.<br>**Full (AoANet)**: Two people standing on a tennis court holding tennis rackets. | GT1: Two young sexy women holding tennis racquets and tennis balls.<br>GT2: A pair of young women hold tennis balls and rackets.<br>GT3: Two women on a tennis court with tennis rackets and balls.<br>GT4: Two female tennis players standing with their rackets.<br>GT5: Two women with tennis rackets and balls are standing. |

Figure 10: More examples.



\end{document}